\documentclass[journal]{IEEEtran}
\usepackage{llncsdoc}
\usepackage{graphicx}
\usepackage{amssymb}
\usepackage{amsmath}
\usepackage{booktabs}
\usepackage{hyperref}

\usepackage{todonotes}

\newcommand\figref{Fig.~\ref}

\newcommand{\beq}{\begin{equation}}
\newcommand{\eeq}{\end{equation}}
\newcommand{\tr}[1]{{#1}^\top}
\newcommand{\vect}[1]{\mathbf{#1}}

\newcommand{\mr}[1]{\mathrm{#1}}

\newcommand{\Real}{\mathbb{R}}

\newcommand{\rref}{{\mathrm{ref}}}

\usepackage{scalerel,stackengine}
\stackMath
\newcommand\reallywidehat[1]{%
\savestack{\tmpbox}{\stretchto{%
  \scaleto{%
    \scalerel*[\widthof{\ensuremath{#1}}]{\kern-.6pt\bigwedge\kern-.6pt}%
    {\rule[-\textheight/2]{1ex}{\textheight}}
  }{\textheight}%
}{0.5ex}}%
\stackon[1pt]{#1}{\tmpbox}%
}

\newcommand{\GG}{\mathcal{G}}
\newcommand{\VV}{\mathcal{V}}
\newcommand{\EE}{\mathcal{E}}
\newcommand{\NN}{\mathcal{N}}
\newcommand{\LLambda}{\mathbf{\Lambda}}
\newcommand{\TTheta}{\boldsymbol{\theta}}
\newcommand{\SSigma}{\mathbf{\Sigma}}
\newcommand{\LL}{\vect{L}}
\newcommand{\II}{\vect{I}}
\newcommand{\uu}{\vect{u}}

\newcommand{\uhat}{\reallywidehat{\vect{u}}}
\newcommand{\xx}{\vect{x}}

\newcommand{\sss}{\vect{s}}

\newcommand{\UU}{\vect{U}}

\newcommand{\YY}{\vect{Y}}

\newcommand{\SSS}{\vect{S}}
\newcommand{\Uhat}{{\reallywidehat{\vect{U}}}}
\newcommand{\Utilde}{\widetilde{\vect{U}}}
\newcommand{\utilde}{\widetilde{\vect{u}}}
\newcommand{\DD}{\vect{D}}
\newcommand{\RR}{\vect{R}}
\newcommand{\AAA}{\vect{A}}
\newcommand{\mmu}{\boldsymbol{\mu}}

\DeclareMathOperator*{\argmin}{arg\,min}

\begin{document}


\newcommand*{\textred}{\textcolor{red}}
\newcommand*{\textblue}{\textcolor{blue}}
\newcommand*{\textgreen}{\textcolor{green}}






 
\title{Learnable Pooling in Graph Convolution Networks for Brain Surface Analysis}


%


%

\author{Karthik Gopinath$^*$,
        Christian Desrosiers, and 
        Herve Lombaert
\thanks{All authors are with ETS Montreal, Canada. Corresponding author: K. Gopinath. \textbf{Email:}~{karthik.gopinath.1@etsmtl.net}} }
%


\maketitle

\begin{abstract}


Brain surface analysis is essential to neuroscience, however, the complex geometry of the brain cortex hinders computational methods for this task. The difficulty arises from a discrepancy between 3D imaging data, which is represented in \textit{Euclidean} space, and the \textit{non-Euclidean} geometry of the highly-convoluted brain surface. Recent advances in machine learning have enabled the use of neural networks for non-Euclidean spaces. These facilitate the learning of surface data, yet pooling strategies often remain constrained to a single fixed-graph. This paper proposes a new learnable graph pooling method for processing multiple surface-valued data to output subject-based information. The proposed method innovates by learning an intrinsic aggregation of graph nodes based on graph spectral embedding. We illustrate the advantages of our approach with in-depth experiments on two large-scale benchmark datasets. The flexibility of the pooling strategy is evaluated on four different prediction tasks, namely, subject-sex classification, regression of cortical region sizes, classification of Alzheimer's disease stages, and brain age regression. Our experiments demonstrate the superiority of our learnable pooling approach compared to other pooling techniques for graph convolution networks, with results improving the state-of-the-art in brain surface analysis.

\end{abstract}

\section{Introduction}

\begin{figure}[t!]
  \centering
  \mbox{
    \includegraphics[width=0.85\linewidth]{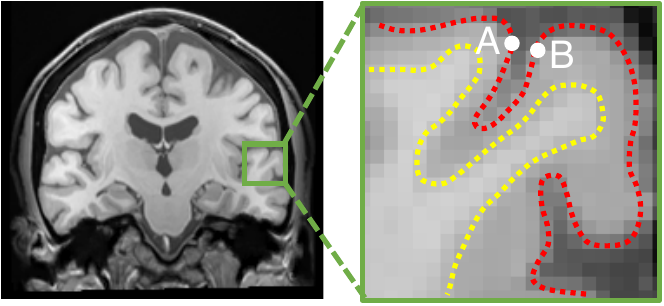}
}
  \caption{Complex geometry of the cerebral cortex. As illustrated, two nearby points in the volume may in fact be far apart on the cortical surface.}
  \label{brain_surf}
\end{figure}

Brain surface analysis plays a crucial role in understanding the mechanisms of perception and cognition in humans \cite{Arbabshirani2017Single}. However, the complex geometry of the brain surface, comprised of intricate folding patterns, poses considerable challenges in neuroscience. 
Notably, brain imaging data, for instance acquired by magnetic resonance imaging, typically comes in 3D, a \textit{Euclidean} space, while its analysis often focuses on the thin surface of the brain, a \textit{non-Euclidean} space. This fundamental difference between the domains of acquisition and analysis, coupled with the geometrical complexity of brain surfaces, severely hinders computational approaches for brain surface analysis.
As an illustration, neighboring 3D voxels in a neuroimage may in fact represent points that are far apart on the brain surface, as shown on Fig.~\ref{brain_surf}. 
%
To alleviate this problem, popular surface-based methods \cite{fischl2004automatically,Yeo2010Spherical} often simplify the geometry of the brain, for instance, by mapping the surface to a sphere. This process is, however, computationally expensive. For example, the widely-used surface analysis pipeline of FreeSurfer \cite{fischl2004automatically} requires several hours to inflate the cortical surface to a sphere, match it to an atlas and finally perform a cortical analysis. 
The geometry of brain surfaces similarly complicates other conventional approaches for brain analysis, such as those based on diffeomorphic transformations \cite{Glaunes2004Diffeomorphic} or on spherical harmonics \cite{Styner2006Framework}. 

A key application of brain surface analysis is detecting and tracking the progress of neurodegenerative disorders, such as Alzheimer's disease, which often result in a severe atrophy of brain tissues. Analyzing the geometrical changes of the brain can thus aid in the early diagnosis of such conditions. Initial work has focused on \textit{Euclidean} 3D data, based for instance on the texture of magnetic resonance images \cite{vemuri2008alzheimer, freeborough1998mr}, in order to differentiate Alzheimer's disease from normal aging. 
While volumetric approaches has shown relevance in detecting global changes in a Euclidean space \cite{Arbabshirani2017Single}, surface-based methods \cite{fischl2004automatically,Yeo2010Spherical,Glaunes2004Diffeomorphic,Styner2006Framework} are more adequate for analyzing data on brain surfaces. 
For instance, the analysis of shape abnormalities on brain surfaces has improved the prediction of Alzheimer's disease \cite{tang2014shape} or the identification of stages in the disease \cite{oliveira2010use}. 
%
Nevertheless, all these studies has focused on pre-established measurements of brain surface information. 
%
This paper proposes to learn and exploit the organizational structure of surface data in order to improve the prediction tasks that use data on highly complex surfaces. 


\subsection{Related work}

  
    

Current machine learning approaches have achieved state-of-the-art performance in a broad range of computer vision and medical imaging applications. In particular, deep learning architectures, such as convolutional neural networks (CNNs) \cite{Lecun1998Gradientbased}, offer higher accuracy and speed over traditional approaches for image analysis. In neuroimaging, CNNs are now widely used for various segmentation \cite{ronneberger2015u} and classification \cite{wachinger2017deepnat} problems, with architectures tailored for the target task and the available imaging data. For example, various architectures have been proposed to exploit volumetric data  \cite{Zhang2011ODVBA,Hua2013Unbiased,dolz20173d,Kamnitsas2017Efficient}. A fundamental limitation of these models, however, is their restriction to data lying on a fixed Euclidean grid representing pixels or voxels. This restricted representation induces ambiguity when exploiting complex geometries, such as in brain surfaces, impeding the application of these Euclidean models for brain surface analysis. 

Geometric deep learning \cite{Bronstein2017Geometric} generalizes deep learning models to operate on non-Euclidean domains such as graphs and manifolds. Recent advances in this field, in particular, in graph convolution networks (GCNs), have enabled convolution operations over graphs by exploiting spectral analysis, where convolutions translate into multiplications in a Fourier space \cite{Bruna2014Spectral,Kipf2017SemiSupervised,Defferrard2016Convolutional,monti2017geometric}. In such models, convolutions are manipulated with eigenfunctions of graph Laplacian operators \cite{Xu2018SpiderCNN}, which can be approximated with Chebyshev \cite{Defferrard2016Convolutional} or Cayley polynomials \cite{Levie2018CayleyNets}. These learned convolution filters are expressed in terms of mixtures of Gaussians \cite{monti2017geometric} or splines \cite{Fey2018SplineCNN}. Despite their advantages over standard CNNs, these models are, however, limited to a fixed graph structure and thereby not suitable for brain imaging applications involving a population of subjects. Indeed, brain surfaces have varying geometries with a different number of nodes and a distinct connectivity across meshes. This variability poses computational challenges, for example, arising from the fact that the values of a Laplacian eigenfunction can drastically differ between brains with different surface geometries \cite{Ovsjanikov2012Functional}. To this effect, a learned synchronization can correct for differences in eigenfunctions \cite{Yi2017SyncSpecCNN}. An alignment of eigenbases \cite{Lombaert2015Brain} similarly provides a common parameterization of brain surfaces. Such aligned eigenbases enabled the direct learning of surface data across multiple brain geometries \cite{Gopinath2019Graph}. Nevertheless, these types of GCNs are limited to a \textit{fixed} graph structure, for instance, with the same number of nodes. Standard pooling strategies rely in fact on such consistency of graph structures. Currently, heuristics are often used to mimic a max-pooling strategy in GCNs \cite{Bruna2014Spectral,Defferrard2016Convolutional,Dhillon2007Weighted}. They include varying the number of feature dimensions across layers \cite{Bruna2014Spectral} while retaining fixed layer sizes, or relying on partition methods, for instance, based binary trees \cite{Defferrard2016Convolutional} or Graclus clustering \cite{Dhillon2007Weighted} to coarsen the initial graph. However, these strategies are mainly used for point-wise operations in fixed-size graphs \cite{monti2017geometric}, such as node classification \cite{Parisot2017Spectral}, and do not apply to the task of subject classification when the geometry varies across subjects.

A few recent studies \cite{wang2018local,Ying2018Hierarchical} have attempted to tackle the problem of graph classification in GCNs by incorporating adaptive pooling modules in the network. For instance, \cite{wang2018local} performs a hierarchical clustering of nodes using their spectral coordinates, with a subsequent pooling of node features within each cluster. While this approach handles varying graph structures, clusters are defined only on node proximity in the embedding space, without considering its values. Consequently, this unsupervised pooling strategy may not be optimal for the classification or regression task at hand. 
More recently, a differential pooling technique \cite{Ying2018Hierarchical} splits the network in two separate paths, one for computing latent features for each node of the input graph and another for predicting the node clusters by which features are aggregated. 
%
%
This approach ignores, however, the intrinsic localization of nodes within the graph, which is sought when the geometry is highly curved, such as, in particular, brain surfaces.

\begin{figure*}[t]
  \centering  
  
    \includegraphics[width=\textwidth]{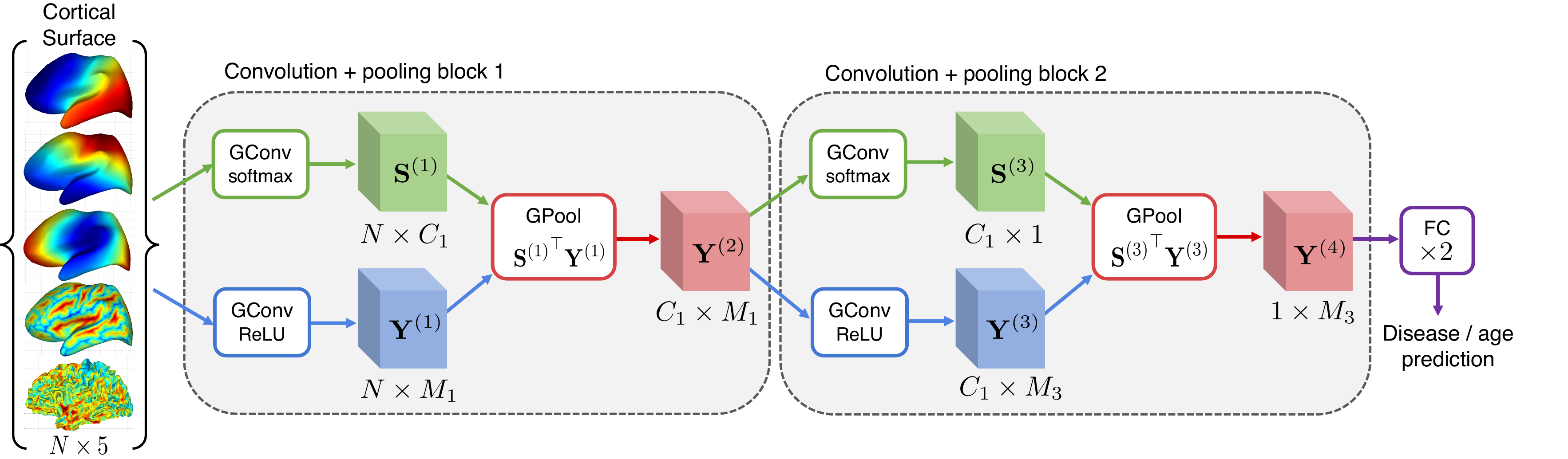}
    
  \caption{An overview of the proposed graph convolution network for subject-specific cortical surface analysis.}
  \label{reg_arch}
\end{figure*}


\subsection{Contributions}


This paper proposes a novel method based on GCNs for classification and regression of surface graphs. Our method includes a learnable pooling strategy which predicts optimal node clusters for each  input graph, and thus can handle graphs with varying number of nodes or connectivity. This adaptive pooling technique is applied recursively to obtain a fixed-size representation, which is then used for predicting a target classification or regression value. Our method also leverages spectral embedding techniques for surface graphs \cite{Lombaert2015Brain}, offering a more powerful representation of complex surfaces like the brain cortex. This contrasts with the differential pooling approach in \cite{Ying2018Hierarchical}, where nodes lack intrinsic localization within the graph.

We illustrate our approach on the challenging tasks of brain surface classification and regression using the well-known Mindboggle and ADNI datasets. We first compare our learnable pooling strategy to other pooling techniques for GCNs, and study the effect of input graph size (i.e., surface mesh resolution) on performance, by considering the problem of subject sex classification\footnote{As in most studies, we use the term \emph{sex} instead of \emph{gender} to designate biological differences between male and female subjects.}. The ability of our pooling strategy to learn important node clusters in a supervised manner is highlighted by the relationship between these clusters and prominent anatomical regions. To further validate the regions learned by our network, we use it to predict the size of cortical regions as defined by a standard parcellation atlas. 
Our model is also tested on cortical surface data from the ADNI dataset to \emph{i}) discriminate between control subjects and subjects suffering from different stages of Alzheimer's, and \emph{ii}) regress the brain age of subjects. We choose the largest dataset ADNI \cite{Esther2015Standardized} as it provides manual labels for subject brain age and three stages of Alzheimer's disease. Only using simple cortical measurements such as thickness and sulcul depth, our method achieves a similar performance to the state-of-the-art on the ADNI dataset \cite{Esther2015Standardized}. 


In summary, the major contributions of our work are as follows:
\begin{itemize}\setlength\itemsep{0.4em}
\item A general model for classifying and regressing graphs with varying geometry, which combines a learnable, supervised pooling strategy with the intrinsic (non-Euclidean) localization of nodes via graph spectral embedding. 
\item A first fully-learned model for brain surface analysis contrasting with previous approaches based on predefined cortical features; 
\item An in-depth experimental evaluation on two large-scale benchmark datasets (i.e., Mindboggle and ADNI) and four different prediction tasks (i.e., subject sex classification, cortical region size regression, Alzheimer's disease classification, and brain age regression); 
%
\item State-of-the-art performance for ADNI stages classification and brain age prediction using cortical surface data.
\end{itemize}

This paper represents a significant extension of our previous work in \cite{gopinath2019adaptive}. Specifically, we test our method on another multi-site dataset (i.e., Mindboggle) and explore two additional prediction tasks (i.e., subject-sex classification and cortical region size regression). Results of these new experiments highlight the relationship between learned clusters for these tasks and known cortical regions. This extended study also compares our model against graph pooling techniques relying on unsupervised spectral clustering \cite{wang2018local} and differentiable pooling approaches in Euclidean space \cite{Ying2018Hierarchical}, showing significant advantages compared to these techniques. Last, additional experiments are proposed to show the robustness of our method to surface mesh variability in terms of number of nodes and connectivity. 



\section{Method}

We first describe a general formulation that extends standard convolutions to non-rigid geometries such as surfaces. We then detail our strategy based on graph spectral embeddding to model the intrisic localization of mesh nodes and align them across multiple surfaces. Subsequently, we present our end-to-end learnable pooling strategy for the adaptive clustering of graph nodes. Finally, we provide detailed information on the overall network architecture and training procedure.  



\subsection{Convolutions on non-rigid geometries}
\begin{figure*}[t]
  \centering
  \mbox{
    \shortstack{
    \includegraphics[width=0.35\textwidth]{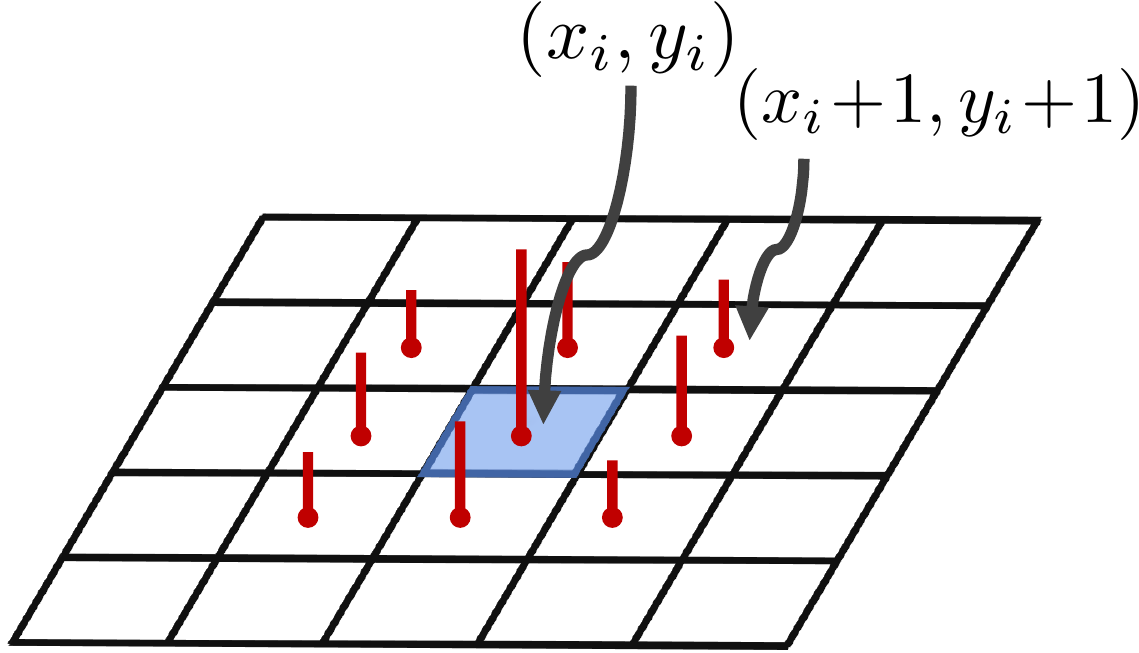}
    \\ \phantom{M}}
    
    \hspace{5mm}
    
    \includegraphics[width=0.41\textwidth]{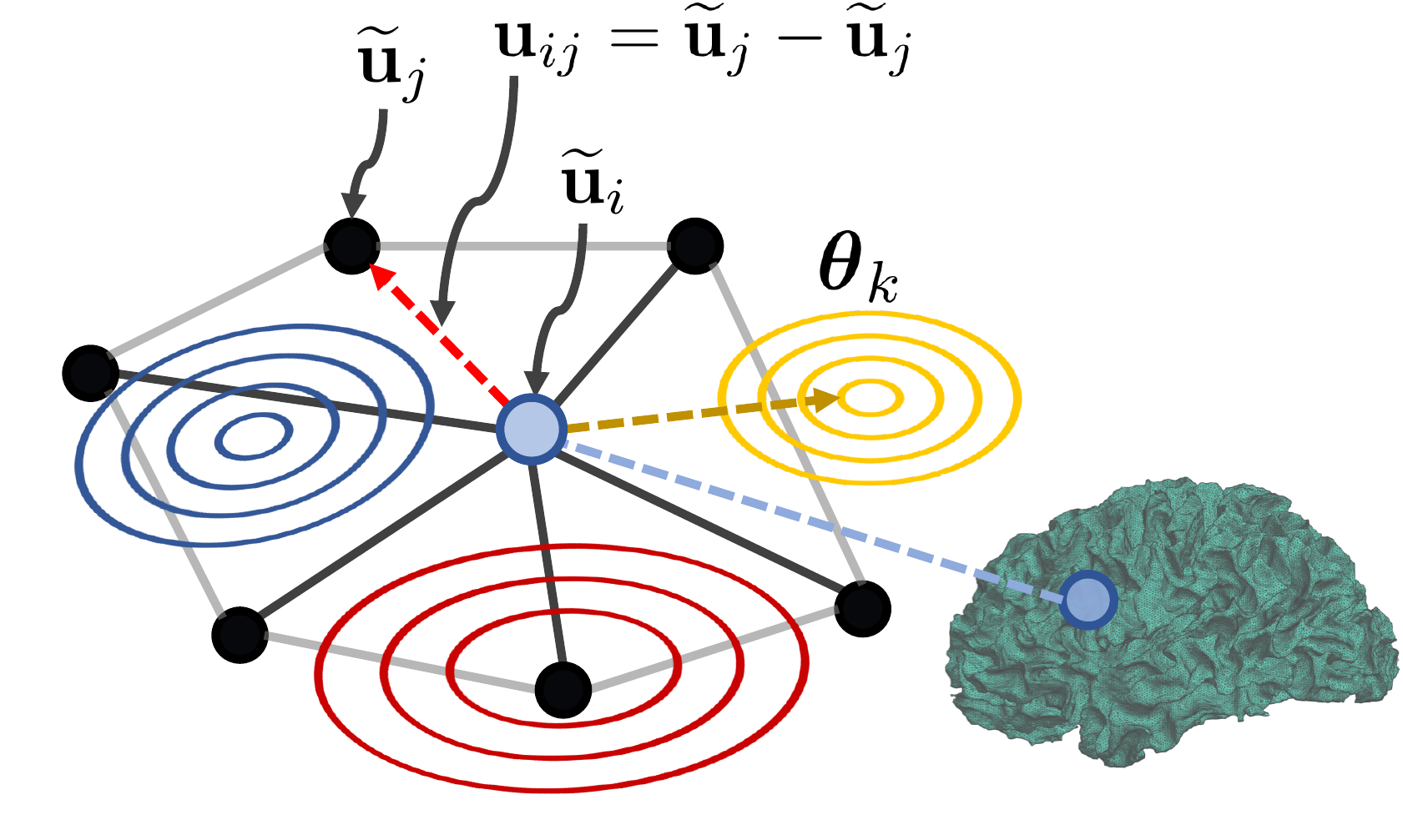}
}
  \caption{Illustration of standard grid-based 2D convolutions (left) and geometric graph convolution (right). The challenge is to exploit kernels on arbitrary graph structures, and to add pooling operations over convolutional layers of graph nodes.}
  \label{grid_vs_graph_conv}
\end{figure*}

In a standard CNN, the input is typically provided as a set of features observed over a regular grid of points like 2D pixels or 3D voxels. This information is then processed using a sequence of layers composed of a convolution operation followed by a non-linear activation function like the ReLU. Let $\YY^{(l)} \in \Real^{N \times M_l}$ be the input feature map at convolution layer $l$, such that $y_{iq}^{(l)}$ is the $q$-th feature of the $i$-th input node. The feature map consists of $N$ input nodes with $M_l$ dimensions each. Assuming a 1D grid for simplicity, the output of layer $l$ obtained by a convolution kernel of size $K_l$ is given by $y_{ip}^{(l+1)} = f(z_{ip}^{(l)})$, 
where 
\begin{equation}\label{eq:1D-convolution-z}
    z_{ip}^{(l)} \ = \ \sum_{q=1}^{M_l} \sum_{k=1}^{K_l} w_{pqk}^{(l)} \cdot y_{i+k,\, q}^{(l)} 
            \ + \ b_p^{(l)}. 
\end{equation} 
Here, $w_{pqk}^{(l)}$ are the convolution kernel weights, $b_p^{(l)}$ the weights of the layer, and $f$ the activation function.

For a general surface, points are not necessarily defined on a regular grid and can lie anywhere in a 3D Euclidean space. Such surface can conveniently be represented as a mesh graph $\GG = \{\VV, \EE\}$ where $\VV$ is the set of nodes corresponding to points and $\EE$ is the set of edges between the graph nodes. Given a node $i \in \VV$, we denote as $\NN_i  = \{j \,| \,(i,j) \in \EE\}$ the set of nodes connected to $i$, called neighbors. We extend the concept of convolution to arbitrary graphs using the more general definition of geometric convolution \cite{monti2017geometric,Gopinath2019Graph,Fey2018SplineCNN}:
\begin{equation}\label{gCNN}
    z_{ip}^{(l)} \ = \ 
        \sum_{j \in \NN_i} \sum_{q=1}^{M_l} \sum_{k=1}^{K_l} 
            w_{pqk}^{(l)} \cdot y_{jq}^{(l)} \cdot \varphi_{ij}(\TTheta^{(l)}_k)  \ + \ b_p^{(l)},
\end{equation}
In this extended formulation, $\varphi_{ij}$ is a symmetric kernel parameterized by $\TTheta_k$, which encodes the relative position of neighbor nodes $j$ to a node $i$ when computing the convolution at node $i$. For instance, $\varphi_{ij}$ can be defined as a Gaussian kernel with learnable parameters $\TTheta_k = \{\mmu_k, \SSigma_k\}$ on the local polar coordinate $\uu_{ij} = (\phi_{ij}, \theta_{ij})$ from node $i$ to $j$ \cite{monti2017geometric}:
\begin{equation}\label{eq:kernel}
  \varphi_{ij}(\TTheta_k) \ = \ \exp\big(-\tfrac{1}{2}\tr{(\uu_{ij} - \mmu_k)}\SSigma_k^{-1}(\uu_{ij} - \mmu_k)\big).
\end{equation}
The relationship between conventional and geometrical convolutions is illustrated in Fig. \ref{grid_vs_graph_conv}. Standard convolutions (left) can in fact be seen as a special case of geometric convolutions (right) where nodes are placed on a regular grid and kernels are unit impulses (i.e., spherical Gaussian kernels with zero variance) placed at the grid position of neighbor nodes.

\subsection{Spectral embedding of multiple surface graphs}

A significant limitation of the above geometric convolution model is its inability to process differently-aligned surfaces. Thus, since local coordinates $\uu_{ij}$ are determined using a fixed coordinate system, any rotation or scaling of the surface mesh will produce a different response for a given set of kernels. Moreover, as shown in Fig. \ref{brain_surf}, geometric convolutions in Euclidean space are poorly-suited for complex surfaces like the highly-convoluted brain cortex.

We address these issues using a graph spectral embedding approach. Specifically, we map a surface graph $\GG$ to a low-dimensional subspace using the eigencomponents of its normalized Laplacian $\LL = \II - \DD^{-\frac{1}{2}}\AAA\DD^{-\frac{1}{2}}$, where $\AAA$ is the weighted adjacency matrix and $\DD$ is the diagonal degree matrix with $d_{ii}=\sum_j a_{ij}$. 
Although binary adjacency values could be used in $\AAA$, we instead define the weight between two adjacent nodes as the inverse of their Euclidean distance: $a_{ij} = \big(\|\xx_i - \xx_j\|_2 + \epsilon\big)^{-1}$ where $\epsilon$ is a small constant to avoid a zero-division. 
Denoting as $\UU \LLambda \tr{\UU}$ the eigendecomposition of $\LL$, where $\LLambda$ is the diagonal matrix of real, non-negative eigenvalues, we then compute the normalized spectral coordinates of nodes as the rows of matrix $\Uhat = \UU\LLambda^{-\frac{1}{2}}$. Here, normalized components are scaled proportionally to the \emph{inverse} of their eigenvalues since components with smaller eigenvalues encode more relevant characteristics of the embedded graph \cite{chung1997spectral}. Based on the same principle and as in \cite{Lombaert2015Spectral}, we limit the decomposition to the $d=3$ first smallest non-zero eigenvalues of $\LL$. This allows capturing the important variability of surfaces, while also limiting computationally complexity.  

Because the spectral embedding of $\LL$ is only defined up to an orthogonal transformation (i.e., rotation or flip), we must align the spectral projection of different surface graphs to a common reference $\Uhat^\rref$. Toward this goal, we use an iterative closest point (ICP) method \cite{Lombaert2015Brain} where each node $i \in \VV$ is mapped to its nearest reference node $\pi(i) \in \VV^\rref$ in the embedding space. Denoting as $\uhat_i$ the normalized spectral coordinates of node $i$, the alignment task can be expressed as
\begin{equation}\label{eq:correspondence}
    \argmin_{\pi, \RR} \ \sum_{i=1}^N \big\| \uhat_i \, \RR \,- \,\uhat^\rref_{\pi(i)} \big\|_2^2. 
\end{equation}
Let $\Uhat^\rref_\pi$ be the matrix whose $i$-th row is $\uhat^\rref_{\pi(i)}$. The transformation between corresponding nodes is approximated as
\begin{equation}
    \RR \ = \ \big(\tr{\Uhat} \Uhat\big)
^{-1} \tr{\Uhat} \Uhat^\rref_{\pi} \ = \  
    \LLambda^{\frac{1}{2}} \tr{\UU} \Uhat^\rref_{\pi}.
\end{equation}
We use the aligned spectral embedding $\Utilde = \Uhat\,\RR$ to define the local coordinates corresponding to an edge $(i,j) \in \EE$: $\uu_{ij} = \utilde_j - \utilde_i$. As illustrated in Fig. \ref{grid_vs_graph_conv} (right), and based on Eq. (\ref{gCNN}), the convolution at node $i$ therefore considers kernel responses $\varphi_{ij}(\TTheta^{(l)}_k)$ for neighbor nodes $j$, \emph{relative to} the spectral coordinates of $i$.

\subsection{Learnable pooling for graph convolution networks}

Pooling in standard CNNs is typically carried out by aggregating values inside non-overlapping regions of features maps. In graph convolution networks \cite{Bruna2014Spectral,Kipf2017SemiSupervised,Defferrard2016Convolutional,monti2017geometric}, however, this approach is not applicable for the following reasons. First, nodes are not laid out on a regular grid, which prevents aggregation of features in pre-defined regions. Second, the density of points may spatially vary in the embedding space, hence regions of fixed size or shape are not suitable for graphs with different geometries. Last, and more importantly, input surface graphs may have a different number of nodes, while the output may have a fixed size. This is the case when predicting a fixed number of class probabilities from different brain geometries.

We propose an end-to-end learnable pooling strategy for the subject-specific aggregation of cortical features, inspired by the differential pooling technique of Ying et al. \cite{Ying2018Hierarchical}. Our strategy, shown in Fig. \ref{reg_arch}, splits the network in two separate paths: the first one computing latent features for each node of the input graph and the second predicting the node clusters by which the features are aggregated. The feature encoding path is similar to a conventional CNN, and produces a sequence of convolutional feature maps $\{\YY^{(1)}, \ldots, \YY^{(l)}\}$ with $\YY^{(l)} \in \Real^{N \times M_l}$. The clustering path consists of sequential convolutional blocks, however the activation function of the last block is replaced by a node-wise softmax. The output of this last block, $\SSS \in [0,1]^{N \times C}$, gives for each node $i$ the probability $s_{ic}$ that $i$ belongs to cluster $c$. Pooled features $\YY^{\mr{pool}} \in \Real^{C \times M_l}$ are computed as the expected sum of convolutional features in each cluster, i.e.
\beq\label{eq:pooling}
    \quad\quad y^{\mr{pool}}_{cp} \, = \, \sum_{i=1}^N s_{ic} \cdot y^{(l)}_{ip}, \quad\quad   \YY^{\mr{pool}} \, = \, \tr{\SSS} \YY^{(l)}
\eeq
The processing of aggregated node features, downstream the pooling operation, requires computing a new adjacency matrix $\AAA^{\mr{pool}}$ for the node clusters. Here, we define the adjacency weights between pooling clusters $c$ and $d$ as 
\beq\label{eq:pooling-adj}
    a^{\mr{pool}}_{cd} \, = \, \sum_{i=1}^N \sum_{j=1}^N s_{ic} \cdot s_{jd} \cdot a_{ij}, \quad   \AAA^{\mr{pool}} \, = \, \tr{\SSS} \AAA \SSS.
\eeq
Intuitively, $a^{\mr{pool}}_{cd}$ is the expected number of connected nodes between clusters $c$ and $d$.

As mentioned in \cite{Ying2018Hierarchical}, the bilinear formulation of Eq. (\ref{eq:pooling}) faces a challenging optimization problem with several local minima. For instance, the same output $\YY^{\mr{pool}}$ in Eq (\ref{eq:pooling}) can be obtained by modifying either $\SSS$ or $\YY^{(l)}$. To alleviate this problem and obtain spatially-smooth clusters, we add a Laplacian regularization term to the loss function: 
\beq\label{eq:reg_laplace}
    \mathcal{L}_{\mr{reg}}(\SSS) \ = \ \sum_{i=1}^N \sum_{j=1}^N a_{ij} \cdot \|\sss_i - \sss_j\|^2 \ = \ \mr{tr}(\SSS\LL\tr{\SSS}),
\eeq
where $\sss_i$ denotes the cluster probability vector of node $i$ (i.e., the $i$-th row of $\SSS$). This well-known regularization approach  \cite{belkin2006manifold} penalizes connected nodes to be mapped to different clusters, with penalty proportional to the connection strength.


\subsection{Architecture details}\label{sec:architecture}

Figure \ref{reg_arch} presents the overall architecture of our  graph convolution network. As input, we give the network the cortical surface features $\xx_i$ and aligned spectral coordinates $\utilde_i$ of each node $i$. For computing graph convolutions as in Eq. (\ref{gCNN}), we define the neighbors $\NN_i$ of node $i$ as the $k=5$ nodes nearest to $i$ in the spectral embedding (i.e., the distance between node $i$ and $j$ corresponds to $\|\uu_i - \uu_j\|_2$) plus node $i$ itself. While various features could be considered to model the local geometry of the cortical surface \cite{fischl2004automatically}, we considered sulcal depth and cortical thickness in this work, since the first one helps delineate anatomical brain regions \cite{destrieux2009sulcal} and the latter is related to ageing \cite{sowell2004longitudinal} and neurodegenerative diseases such as Alzheimer's \cite{lerch2004focal}.

The network comprises two cascaded convolution-pooling blocks, followed by two fully-connected (FC) layers. The first block generates an $N \times 8$ feature map and an $N \times 16$ cluster assignment matrix, in two separate paths, and combines them using the pooling formulation of Eq. (\ref{eq:pooling}) to obtain a pooled feature map of $16 \times 8$. In the second block, pooled features are used to produce a $16 \times 16$ map of features, pooled in a single cluster. Hence, the second pooling step acts as an attention module selecting the features of most relevant clusters. The resulting $1 \times 16$ representation is converted to a $1 \times 8$ vector using the first FC layer, and then to a $1 \times \mr{nb. outputs}$ vector with the second FC layer. 

Except for the cluster probabilities and network output, all layers employ the Leaky ReLU \cite{nair2010rectified} 
as activation function: $y_{ip}^{(l)} = \max(0.01 z_{ip}^{(l)}, \,z_{ip}^{(l)})$. Moreover, for the graph convolution kernel $\varphi_{ij}$ of Eq. (\ref{gCNN}), we used the B-spline kernels proposed by Fey et al. \cite{Fey2018SplineCNN}. Compared to Gaussian kernels \cite{monti2017geometric}, this kernel has the advantage of making computation time independent from the kernel size. 

For training, the loss function combines the output prediction loss and cluster regularization loss on the convolution-pooling block: 
\begin{equation}
    \mathcal{L}(\TTheta) = \mathcal{L}_\mr{out}(\TTheta) + \alpha\mathcal{L}_\mr{reg}\big(\SSS^{(1)}(\TTheta)\big),
\end{equation} 
where $\alpha$ is a parameter controlling the amount of regularization. For classification tasks (i.e., disease prediction), $\mathcal{L}_\mr{out}$ is set as the cross-entropy between one-hot encoded ground-truth labels and output class probabilities. In the case of regression (i.e., brain age prediction), we use mean squared error (MSE) for this loss. Network parameters are optimized with stochastic gradient descent (SGD) using the Adam optimizer. Experiments were carried out on an i7 desktop computer with 16GB of RAM and a Nvidia Titan X GPU. The model takes less than a second for disease classification or age regression.


\section{Experiments and results}

We validate our method on two large-scale, publicly-available datasets: Mindboggle-101 \cite{Klein2017Mindboggling} and ADNI1 \cite{jack2008alzheimer}. The first one contains T1-weighted MRI from 101 healthy subjects (males: $n$=57, females: $n$=44, age: 20--61 years) collected from 9 different sites. We use this dataset for the tasks of subject-sex classification and cortical region size regression, since both subject-sex labels and manual for 32 cortical parcels are provided with imaging data. The ADNI1 dataset \cite{jack2008alzheimer} is comprised of multi-sequence MRI data from 400 subjects diagnosed with mild cognitive impairment (MCI), 200 subjects with early Alzheimer's disease (AD) and 200 elderly control subjects, obtained from 55 participating sites. 
Both datasets contain brain surface meshes with pointwise cortical thickness and sulcal depth measurements, generated by FreeSurfer\footnote{https://surfer.nmr.mgh.harvard.edu/}. Cortical meshes in these datasets vary from 102K to 185K nodes. 

In a first experiment, we compare the different pooling strategies for graph convolution networks and measure the impact of input graph size on the task of subject classification performance between different pooling methods. We then illustrate the network's ability to learn meaningful node clusters by predicting the size of cortical parcels from an anatomical atlas. Finally, we highlight the advantages of working in the spectral domain on the problems of disease classification (NC \textit{vs} AD, MCI \textit{vs} AD, and NC \textit{vs} MCI) and brain age regression.


\subsection{Comparison of different pooling methods}

\begin{figure*}[t]
  \centering
    \includegraphics[width=.9\linewidth]{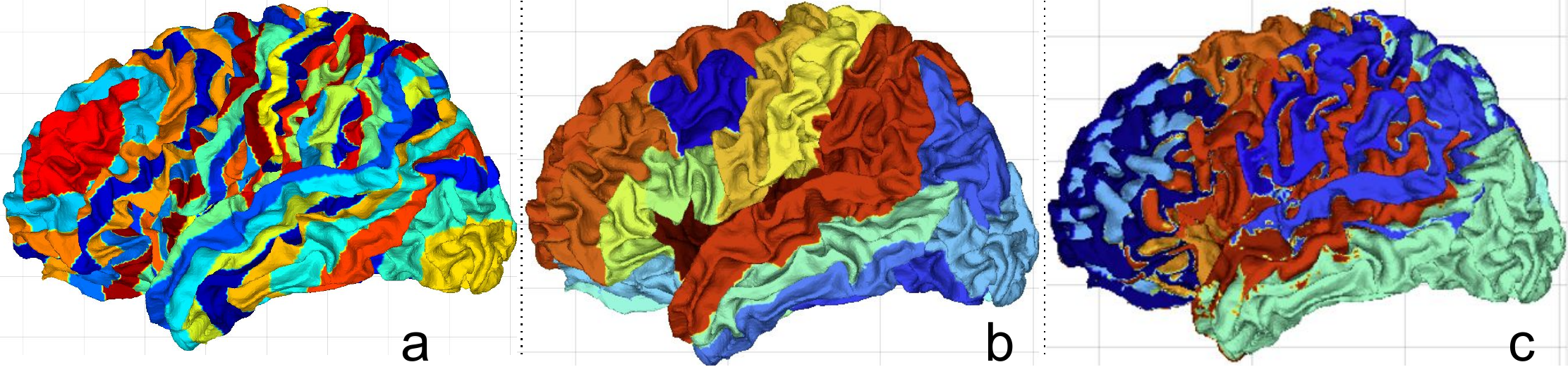}
    
\caption{\textbf{\textbf{Clusters of different pooling methods}}: (a) Clusters obtained by spectral k-means clustering. (b) Fixed clusters computed from a cortical parcel atlas. (c) Clusters learned by our learnable pooling method. Colors on the brain surface represent different regions.}
  \label{spec_parc_our_pool}
\end{figure*}

We compared our learnable pooling strategy against three other pooling techniques applicable to graph convolution networks: 1) taking the global average of feature maps, 2) pooling feature maps in fixed regions computed from a cortical parcel atlas, 3) pooling the same features in regions obtained by applying k-means clustering on the spectral embedding. For all tested methods, we used a network composed of two graph convolution layers followed by two fully-connected layers, as described in Section \ref{sec:architecture}. In the case of global average pooling and fixed parcellation pooling, a single pooling operation is applied after the second graph convolution. For spectral clustering pooling, nodes are grouped after each of the two convolution layers as in our learnable pooling. However, the pooling path of the network is replaced by a static node clustering. We train and test all methods on subject-sex classification Mindboggle dataset with 70-10-20 split for training, validation, and testing. 

\begin{table}[t!]
\label{tabel:diff_pool}
\centering
\setlength{\tabcolsep}{6pt}
\begin{small}
\caption{\textbf{Comparison between different pooling methods:} Average sex classification accuracy, in \%, with standard deviation over the Mindboggle dataset. The first row shows the performance of graph convolution network with global average pooling. The second row shows using fixed manual parcels as regions to pool. The third row shows the performance of spectral clustering pooling method \cite{wang2018local}. Last row indicates the results of our model with learnable pooling.}
\begin{tabular}{lc}
\toprule
\textbf{Pooling methods} & \textbf{Mean $\pm$ Std.} \\ \midrule \midrule
Global Average Pooling & 60.76 $\pm$ 3.62 \\ 
Fixed Parcellation Pooling & 64.59 $\pm$ 7.84 \\ 
Spectral Clustering Pooling \cite{wang2018local} & 67.94 $\pm$ 4.97 \\ 
\midrule
Learnable Pooling (Ours) & 81.33 $\pm$ 5.93 \\ \bottomrule
\end{tabular}
\end{small}
\end{table}

Table~\ref{tabel:diff_pool} summarizes the results of this experiment. We see that global average pooling yields the poorest performance with a mean accuracy of 60.76\%. Using atlas-defined cortical parcels to aggregate features improves accuracy slightly to 64.59\%, suggesting that these parcels are informative of identifying subject sex. Moreover, applying unsupervised clustering on the spectral embedding further increases mean accuracy to 67.94\%, which indicates the benefits of having a hierarchy of non-fixed clusters. However, by learning clusters in a supervised manner from spectral embeddings, our method achieves the outstanding accuracy of 81.33\%, an improvement of 13.39\% over spectral clustering. Figure~\ref{spec_parc_our_pool} gives examples of clusters for the different pooling strategies (except global average pooling which consider all nodes as part of a single cluster). While spectral clustering yields spatially-regular clusters, the distribution of these clusters is arbitrary and does not seem to match known parcels of the cortex (shown in Fig.~\ref{spec_parc_our_pool}b). In contrast, the clusters predicted by our pooling strategy are larger and better align with these known parcels.

\subsection{Impact of input graph size on performance}

The previous experiment considered detailed surface meshes comprised of 102K to 185K nodes, each with a fixed set of edges connecting nodes. In this second experiment, we investigate whether our method is robust to variability in the size of the surface mesh. Toward this goal, we use the same split of the Mindboggle dataset as in the first experiment, and randomly sub-sample the original mesh to 100, 1K, 5K, 10K, 25K, 50K, and 75K nodes. Because convolutions at each node use information from its $k=5$ nearest neighbors, as described in Eq. (\ref{gCNN}), testing multiple sub-sampling with the same number of nodes also assesses the robustness of our model to variations in graph connectivity. We train our model on each of these reduced graph datasets to predict the sex of Mindboggle subjects. 

Table \ref{size_table} gives the classification accuracy for different sizes of training graphs when testing on sub-sampled graphs of the same size or the original full-sized graph. The first case evaluates whether the same accuracy can be achieved with less information at the input of the network, whereas the second case tests if the convolution parameters learned by the network generalize to larger graphs. As expected, classification performance decreases when reducing the size of input graphs, both when testing on sub-sampled graphs and full-sized graphs. When testing on sub-sampled graphs, accuracy drops from 84.21\% while training with full graphs to 55.02\% for graphs with only 100 nodes. However, high accuracy of 81.33\% can be achieved with training graphs of 50K nodes, about half the size of the original graphs. Furthermore, we see that our model trained with moderately-reduced graphs can still perform well on full-sized ones. For instance, the model trained with graphs of 50K nodes achieves an accuracy of 78.94\% when tested on original graphs with about twice this number of nodes.


\begin{table}[ht!]
\centering
\label{size_table}
\setlength{\tabcolsep}{6pt}
\begin{small}
\caption{\textbf{Subject-sex classification performance of our pooling approach on different sub-graphs:} Mean classification accuracy (\%) with standard deviation over multiple test set from the Mindboggle dataset.}
\begin{tabular}{ccc}
\toprule
\textbf{No. of nodes} & \textbf{\begin{tabular}[c]{@{}c@{}}Testing on \\ Sub-sampled graphs\end{tabular}} & \textbf{\begin{tabular}[c]{@{}c@{}}Testing on\\  Full graphs\end{tabular}} \\ \midrule \midrule
100 & 55.02 $\pm$ 13.18 & 52.63 $\pm$ -- \\ 
~1k & 55.98 $\pm$ ~4.25 & 52.63 $\pm$ -- \\ 
~5k & 64.11 $\pm$ ~1.58 & 47.36 $\pm$ -- \\ 
10k & 67.94 $\pm$ ~5.98 & 52.63 $\pm$ -- \\
25k & 71.77 $\pm$ ~4.86 & 73.68 $\pm$ -- \\
50k & 81.33 $\pm$ ~5.94 & 78.94 $\pm$ -- \\ 
75k & 82.30 $\pm$ ~2.66 & 84.21 $\pm$ -- \\ 
\midrule
Full graph & 84.21 $\pm$ -- & 84.21 $\pm$ -- \\ \bottomrule
\end{tabular}
\end{small}
\end{table}

\subsection{Task-specific pooling regions}

\begin{figure*}[t]
  \centering
    \includegraphics[width=\linewidth]{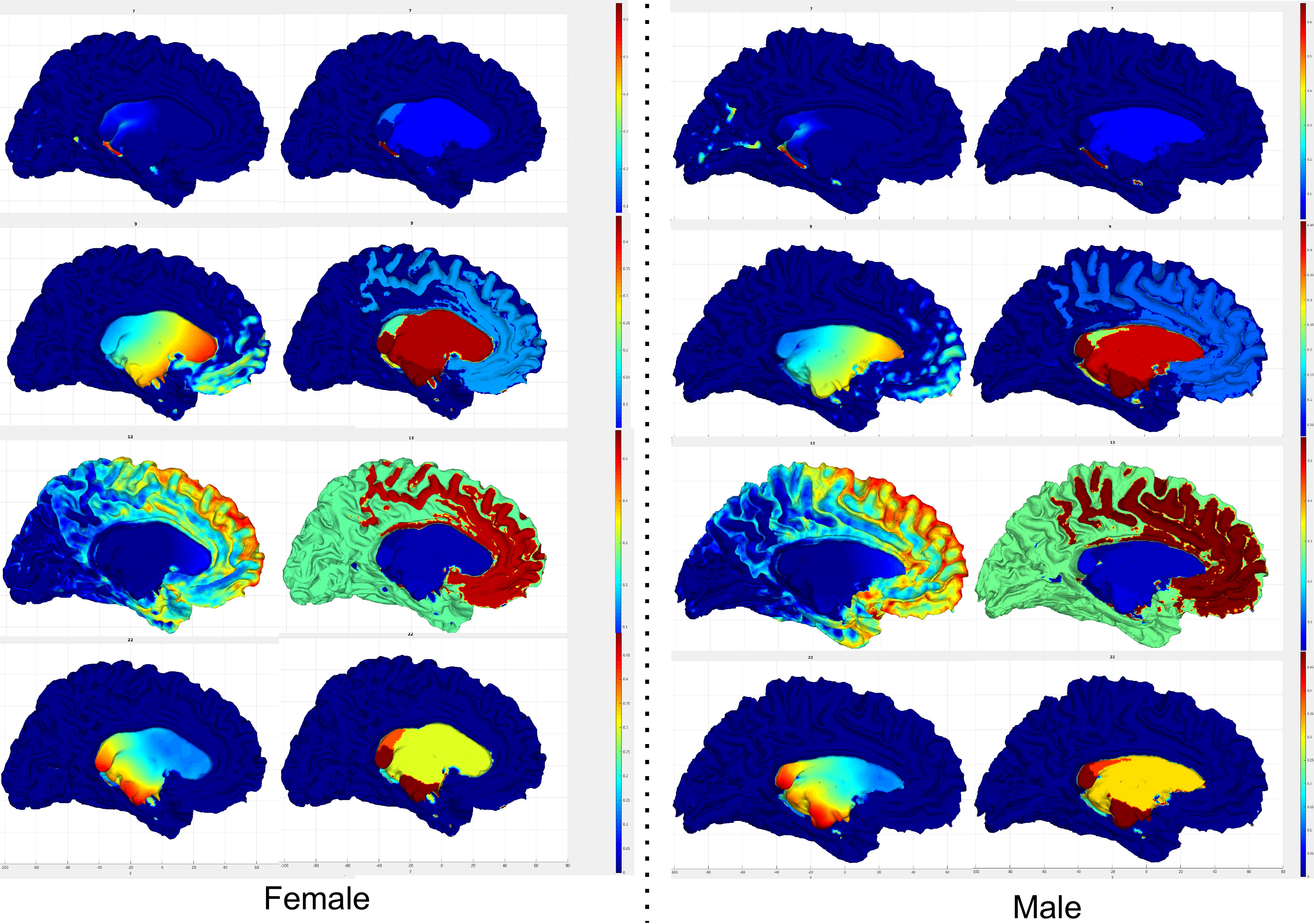}
    
\caption{\textbf{Feature maps and predicted clusters for the task of subject-sex classification}: The first column shows examples of activation maps computed by the embedding path of our network for a female subject. The second column gives the average activation in each predicted cluster for the same subject and feature maps. Third and fourth columns depict the same information for a male subject. 
}
  \label{ml_vs_fml}
\end{figure*}

In this section, we qualitatively and quantitatively evaluate the predicted clusters and feature maps learned by our network. Once more, we consider the task of classifying males \emph{vs.} females subjects from the Mindboggle dataset with the architecture depicted in Fig.~\ref{reg_arch}. 

Figure \ref{ml_vs_fml} shows examples of features and clusters learned by our graph pooling model for a male and a female subject. The first and third columns give the distribution of four different activation maps learned by the network for the two subjects. The mean activation in each predicted cluster for the same subjects is illustrated in the second and fourth columns of the figure. We observe the diversity of depicted clusters, spawning different regions of the brain both on the cortex and around regions of the basal ganglia. Interestingly, several of the learned clusters focus on sub-cortical regions like the hippocampus (first row) and amygdala (last row) which have been linked to sex-related differences in the literature \cite{murphy1996sex}. This illustrates the benefit of learning task-specific clusters in a supervised manner. Additionally, we see that predicted feature maps and clusters in both subjects are similar, demonstrating our model adapts to the specific brain geometry of individual subjects.

We further evaluate the relevance of learned clusters by training the same model to predict the size of 32 anatomical parcels of each brain surface, using labeled data from Mindboggle. This experiments hypothesize that the network should learn clusters which are related to the pre-defined parcels. To do so, we modify the last layer of the architecture in Fig.~\ref{reg_arch} to have 32 outputs, one for the size of each parcel, and change the loss function to mean square error. Adjusted mutual information (AMI) is used to measure the similarity between learned clusters and ground-truth parcels. AMI values range from 0 to 1, a score of 0 corresponding to random clusters and a score of 1 for clusters identical to ground-truth. 

\begin{figure*}[t]
  \centering
    \includegraphics[width=\textwidth]{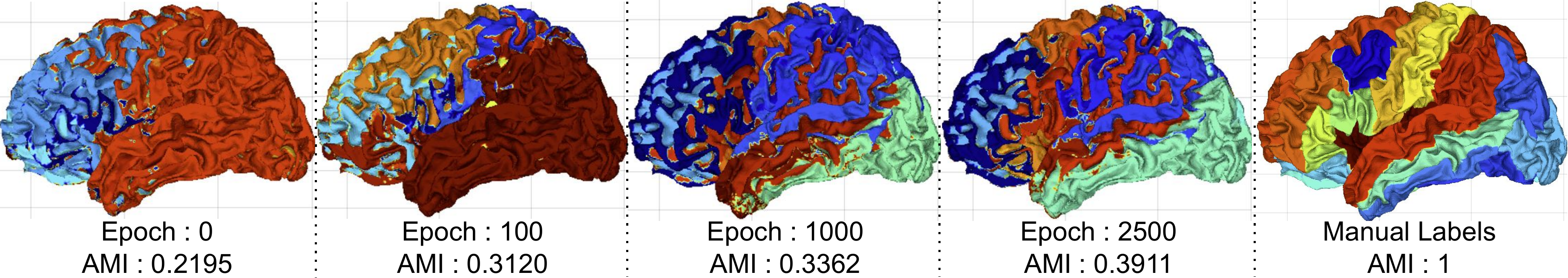}
    
\caption{\textbf{Pooling regions learned during training:} The pooling regions are learned for the model training to regress the size of cortical regions. During initial epochs, random regions are clustered together to aggregate feature maps. A low AMI score indicates this random clustering compared to the ground-truth. After training, the model finally learns to group multiple parcels (cyan) into on cluster pooling region. AMI score increases over epochs indicating task-dependent learning by our model. The last figure shows manual parcels with AMI score of 1 for reference.}
  \label{Clustering_vs_lam}
\end{figure*}

\begin{figure}[t]
  \centering
    \includegraphics[width=\linewidth]{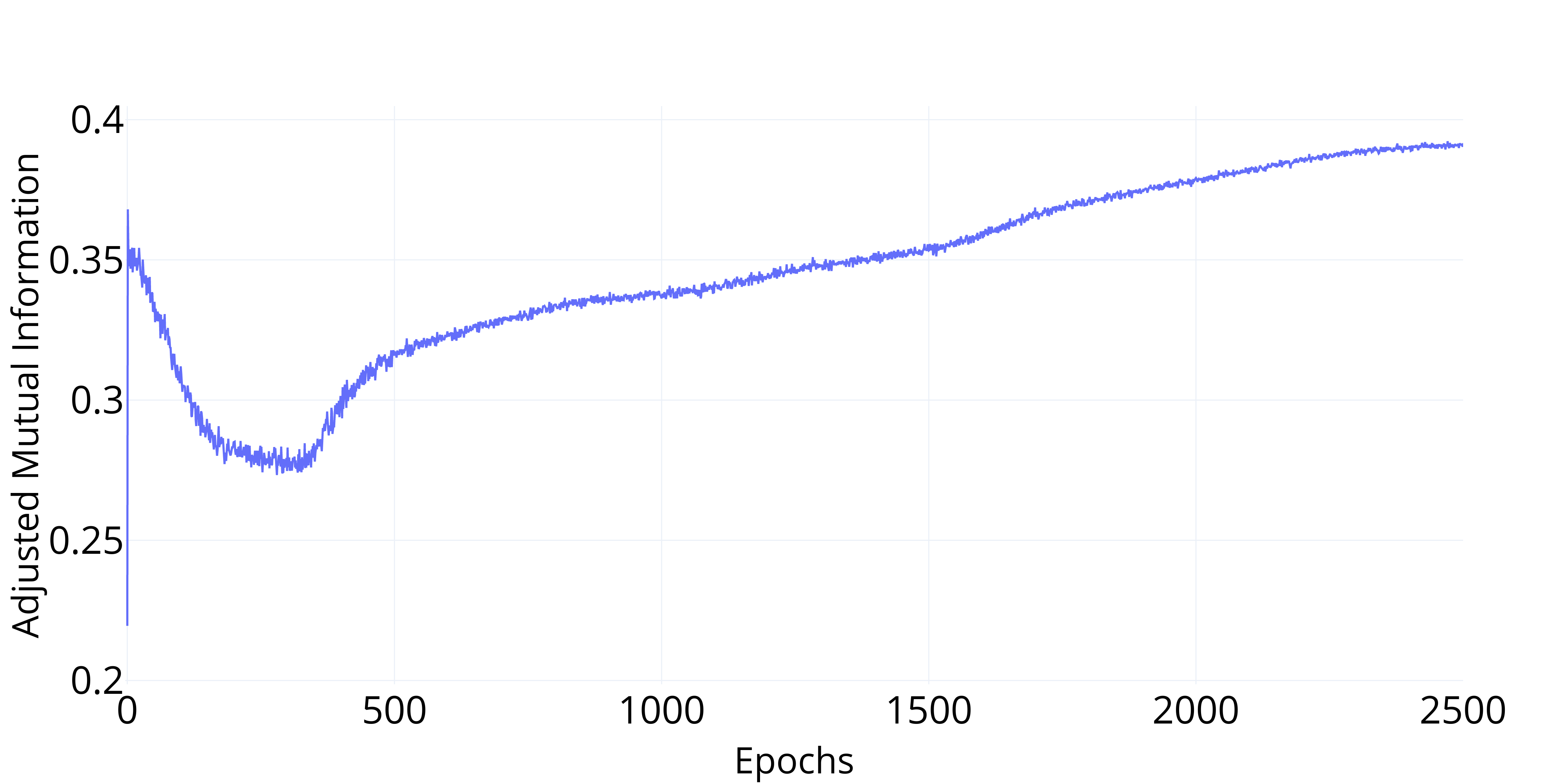}
    
\caption{\textbf{Evolution of AMI score}:  The adjusted mutual information score between the pooling regions and the manual parcels over multiple epochs is shown. A random overlap between learnt pooling regions and parcels is observed at initial epochs. After training, the AMI score increases with the pooling regions corresponding to ground-truth parcels.}
  \label{ep_vs_ami}
\end{figure}

Figure \ref{ep_vs_ami} gives the mean AMI obtained at each training epoch, and examples of predicted clusters at four different epochs are shown in Fig.~\ref{Clustering_vs_lam}. In initial stages of training, the model predicts a small number of clusters corresponding mainly to the components of the spectral embedding (see the network input in Fig.~\ref{reg_arch}). In the first 500 epochs, the AMI score between predicted clusters and ground-truth parcels drops. Then, as training progresses, we observe increasing AMI values and progressively more defined clusters. At the end of training (2500 epochs), the model achieves an AMI score of 0.39. Obtained clusters appear to be a combination of different ground-truth parcels, suggesting that fully-connected layers further help regressing parcel sizes.

\subsection{Disease classification}

In the following experiment, we evaluate our method on the task of classifying subjects from the ADNI dataset as normal control (NC), mild cognitive impairment (MCI) or Alzheimer's disease (AD). Specifically, we consider three different binary classification problems: NC \textit{vs} AD, MCI \textit{vs} AD and NC \textit{vs} MCI. We compare our method against the random forest approach in \cite{ledig2014alzheimer}, which also considers surface-based information from the ADNI dataset. To measure the contribution of the spectral embedding in our method, we also evaluate our model trained with only cortical thickness and sulcal depth as input. The same random split of 70-10-20 is employed for all three models.

The classification performance of tested models is reported in Table~\ref{tbl_acc}. We see that our method outperforms the random forest approach of \cite{ledig2014alzheimer} on all three classification problems. Relative to this approach, the proposed method yields mean accuracy improvements between 7.79\% and 11.92\%. A significant gain in performance is also observed when comparing against the same method trained without spectral node coordinates. This is particularly notable for NC \textit{vs} MCI, where adding spectral coordinates increases the mean accuracy by 13.33\%. Note that we have also tried giving the network original $(x,y,z)$ coordinates of mesh nodes, however this led to worse results. This illustrates the advantage of using intrinsic node localization when processing surface data.

\begin{table*}[htb]
\centering
\setlength{\tabcolsep}{6pt}
\caption{\textbf{Evaluation of the proposed work}: Average accuracy of disease classification (\%), with standard deviation over the complete ADNI dataset. First row is a random forest with multiple cortical-based features \cite{ledig2014alzheimer}. Second row is our graph convolution model without geometrical information (spectral node coordinates). Last row indicates the results of our model with this information.}\label{tbl_acc}
\begin{small}
\begin{tabular}{lcccc}
\toprule
\textbf{Input} & \textbf{NC \textit{vs} AD} & \textbf{MCI \textit{vs} AD} & \textbf{NC \textit{vs} MCI}  \\ \midrule\midrule
Random forest (Cortical-based) \cite{ledig2014alzheimer} & 80 $\pm$ 5        & 65 $\pm$    6 &   63 $\pm$ 4   \\ 
\midrule
\textbf{Ours} (Thickness + Depth)                   & 76.00 $\pm$  6.06      & 74.03 $\pm$  8.63   &   63.71 $\pm$ 5.72  \\ 
\textbf{Ours} (Spectral + Thickness + Depth) & 89.33 $\pm$ 4.30       & 76.92 $\pm$ 4.78     &   70.79 $\pm$ 6.40 \\ \bottomrule
\end{tabular}
\end{small}
\end{table*}


    

\subsection{Brain age prediction}

The last experiment demonstrates our method in a regression problem where the age of NC subjects of the ADNI dataset is predicted using pointwise surface-based measurements. In this case, the network outputs a single value, and MSE is used as a loss function. Once more, we test our method trained with or without spectral node coordinates as input. Moreover, to evaluate brain age prediction as a potential imaging biomarker for Alzheimer's, we also measure the prediction accuracy of our model on AD test subjects.  

\begin{figure*}[t]
  \centering
  \mbox{
    \includegraphics[height=0.3\textwidth]
    {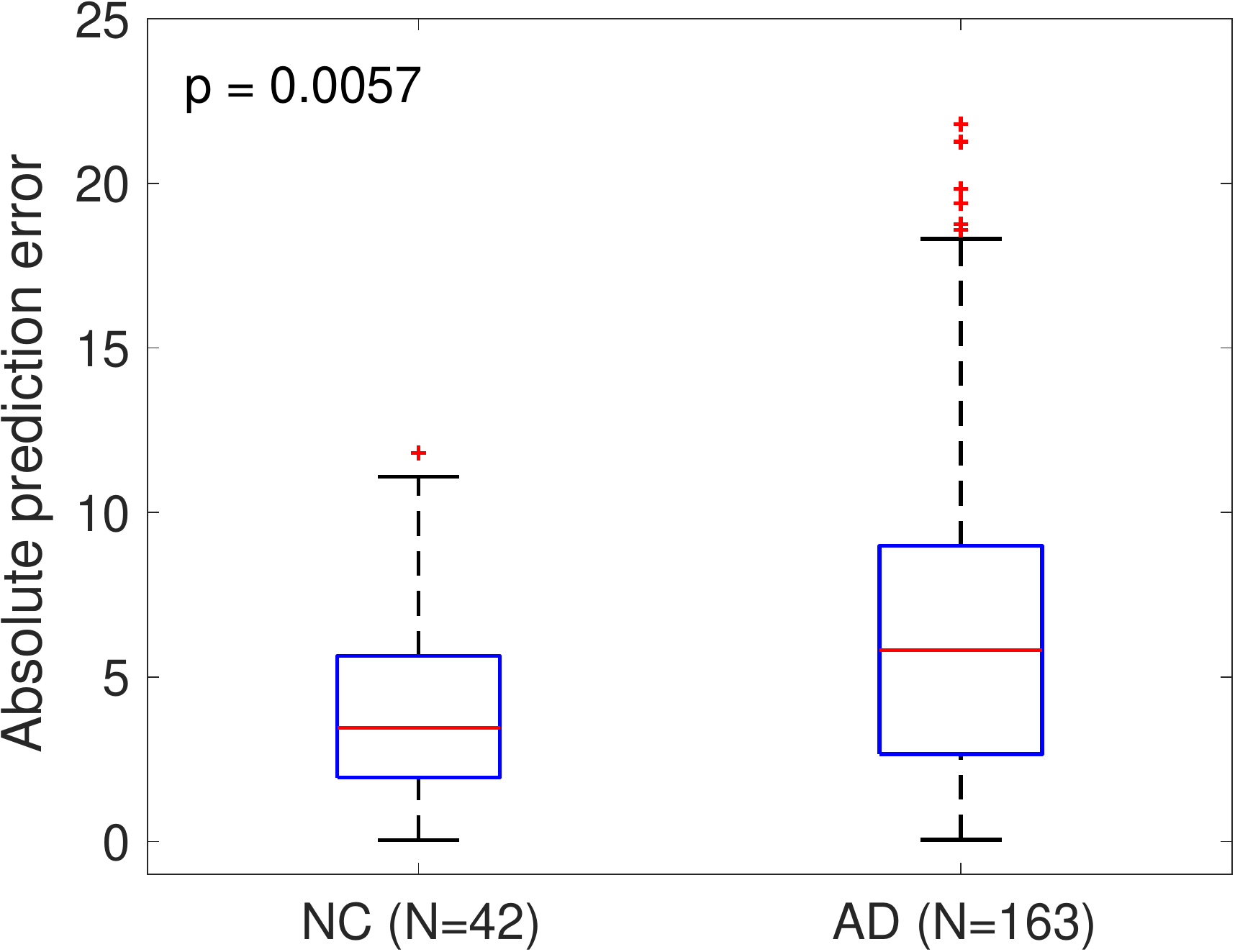}
    
    \hspace{10mm}
    
    \includegraphics[height=0.3\textwidth]{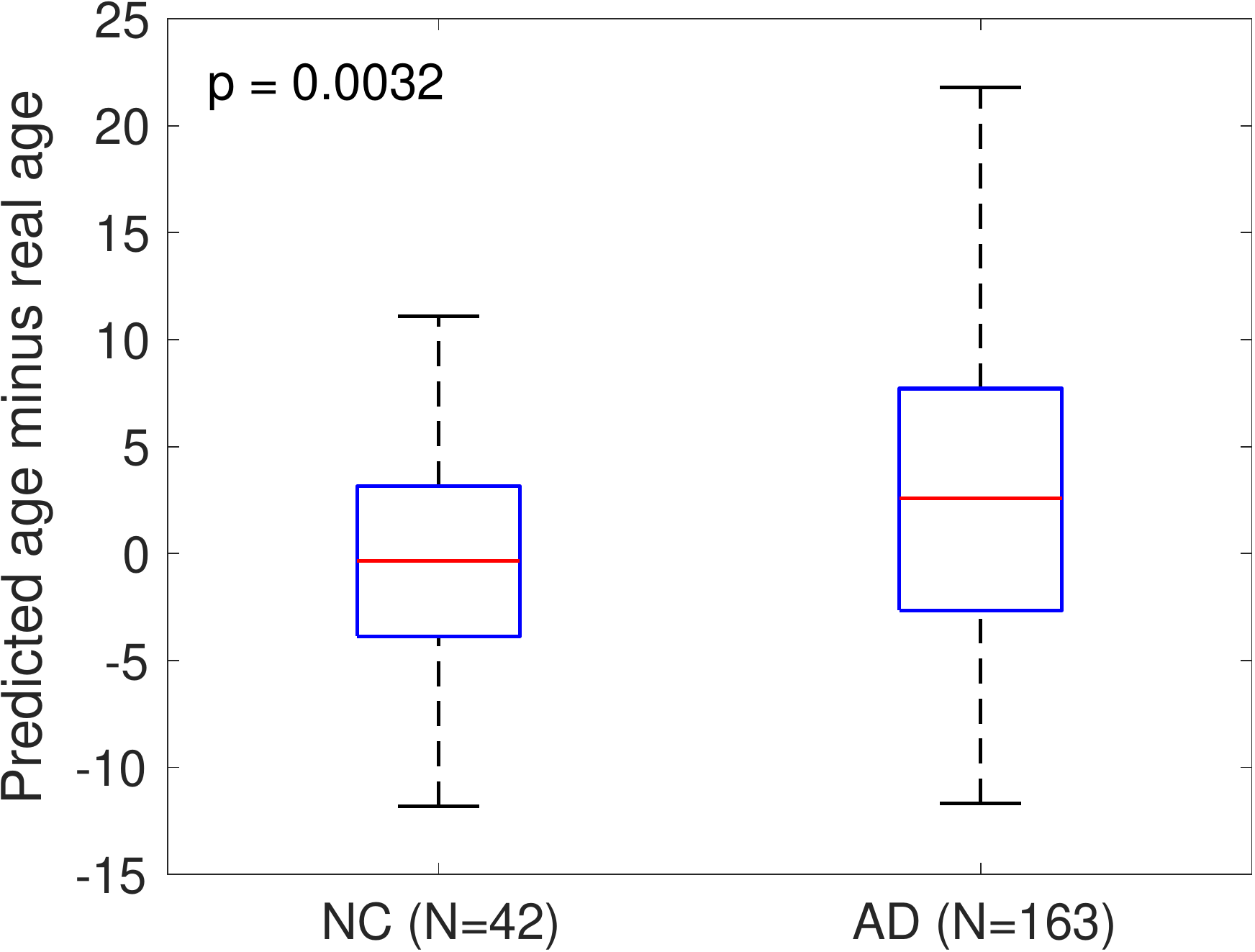}
    }
  \caption{\textbf{Distribution of absolute prediction error} (left) and predicted minus real age (right), for NC and AD test subjects. Our learnable pooling strategy yielded graph models that could correctly capture age discrepancies between real and geometry-based ages, as expected between subjects with NC and AD.}
  \label{brain_age}
\end{figure*}

Results of this experiments are summarized in Fig.~\ref{brain_age}, which gives the distribution of mean absolute error (MAE) and predicted age minus real age for NC subjects and AD subjects. When testing on NC subjects, our method achieves an MAE of 4.35 $\pm$ 3.19 years, which is comparable with results in the literature. As expected, a higher MAE of 6.80 $\pm$ 6 years is obtained for AD subjects, since the symptoms of early Alzheimer's are similar to premature brain aging. The brain age, calculated as the predicted age minus the real age, shows a statistically significant difference with a p-value of 0.0032. This value suggests the potential application of brain age prediction as a biomarker for AD.



\section{Conclusion} 
\label{sec:diss_con}




We presented a novel strategy that enables pooling operations on arbitrary graph structures. Our experiments explored four different applications. In a first experiment, we compared different pooling techniques for graph convolution networks on the subject-sex classification task. A simple global average pooling failed to capture geometric information from consecutive layers, yielding a low performance of 60\%. In comparison to employing fixed pooling regions or learning these regions with unsupervised clustering, our learnable pooling strategy offers significantly higher accuracy.

The second experiments involve assessing the effect of graph size on the performance of subject-sex classification. Results showed that small graphs lack information to capture the complete geometry of surfaces. However, reducing the size of the graph by 25\% up to 75K node does not affect the performance of our model, while improving memory and computational requirements.

The third experiment explored the relationship between learned features and anatomy. The visualization of activation maps and clusters in the network revealed diversity in terms of brain regions. Several learned clusters highlighted essential regions of the basal ganglia, such as the hippocampus and amygdala, which are associated with sex-related differences in the literature. Further evaluation of this result was obtained with an experiment to regress the size of cortical parcels. As expected, the trained model learns pooling regions similar to the manually annotated parcels.

The fourth experiment focused on predicted stages of Alzheimer's disease from surface data, including cortical thickness and sulcal depth. Our results showed that pointwise surface values could be efficiently aggregated into a fixed number of class probabilities using the proposed network architecture. Compared to another approach exploiting surface-based features \cite{ledig2014alzheimer}, our method achieved significant improvements ranging from 7\% to 11\%. This performance gain is mainly due to including spectral coordinates of graph nodes as input to the network, demonstrating the importance of intrinsic node localization.

In a final experiment, the age of ADNI subjects was predicted using pointwise surface data. Results showed our method to provide comparable results as previous approaches in the literature, although only surface-based information is used in our method. As expected, subjects with Alzheimer's have higher discrepancies than subjects with normal cognition (\figref{brain_age}). The potential of the proposed method as an imaging biomarker for AD could be evaluated in a future study.  

To summarize, our pooling strategy enables the exploration of a new family of architectures for graph convolution neural networks. The method exploits the spectral embeddings of graph nodes in order to learn spatially representative pooling patterns across the network layers. However, the proposed method depends on having datasets of comparable brain geometries. The spectral decomposition of graph Laplacian, indeed, assumes that shapes are topologically equivalent. Heterogeneity in holes and cuts in datasets of surfaces remains challenging to exploit since they may produce incompatible sets of Laplacian eigenvectors. This method is consequently inadequate for applications where significant geometrical changes exist, such as when tumors are ablated. It would be interesting to incorporate up-sampling to predict node level output measurements. A new set of methods working on multi-scale would assist brain surface data. Nevertheless, our proposed pooling strategy remains highly-relevant for a wide range of applications where surface data needs to be pooled sequentially in layers from full-size surface-valued vectors to single whole-subject characteristics.

\medskip
\noindent
\textbf{Acknowledgment} -- This work is supported by the Research Council of Canada (NSERC), NVIDIA Corporation with the donation of a Titan Xp GPU. Data were obtained from the Alzheimer's Disease Neuroimaging Initiative (ADNI) database. 


%
%

\addtolength{\textheight}{-5cm}
\bibliographystyle{splncs}
\bibliography{Reference}

\end{document}